\documentclass{article} 
\usepackage{iclr2020_conference,times}

\usepackage{amsmath,amsfonts,bm}

\def\eqref#1{equation~\ref{#1}}

\def\1{\bm{1}}

\DeclareMathAlphabet{\mathsfit}{\encodingdefault}{\sfdefault}{m}{sl}
\SetMathAlphabet{\mathsfit}{bold}{\encodingdefault}{\sfdefault}{bx}{n}

\newcommand{\E}{\mathbb{E}}

\usepackage{amsmath,amssymb,tablefootnote}

\usepackage{hyperref}
\usepackage{url}

\def \E {\mathbb{E}}
\def \bw {\mathbf{w}}

\title{Revisiting Ensembles in an Adversarial Context: Improving Natural Accuracy}
\author{Aditya Saligrama \\ MIT PRIMES \\ \texttt{saligrama@csail.mit.edu} \And Guillaume Leclerc \\ MIT \\ \texttt{leclerc@mit.edu}}

\iclrfinalcopy 
\begin{document}

\maketitle

\begin{abstract}
A necessary characteristic for the deployment of deep learning models in real world applications is resistance to small adversarial perturbations while maintaining accuracy on non-malicious inputs. While robust training provides models that exhibit better adversarial accuracy than standard models, there is still a significant gap in natural accuracy between robust and non-robust models which we aim to bridge. We consider a number of ensemble methods designed to mitigate this performance difference. Our key insight is that model trained to withstand small attacks, when ensembled, can often withstand significantly larger attacks, and this concept can in turn be leveraged to optimize natural accuracy. We consider two schemes, one that combines predictions from several randomly initialized robust models, and the other that fuses features from robust and standard models.
\end{abstract}

\vspace{-1mm}
\section{Introduction}
\vspace{-1mm}
%
%
Adversarial machine learning and its implications \citep{szegedy2014intriguing} on security of neural network decisions has been subjected to significant attention. Recent research has pioneered approaches to defending against adversarial inputs, only to be subsequently confronted with strategies that invalidate these defenses \citep{athalye2018obfuscated, carlini2017adversarial}. 

\citet{madry2018towards} developed a robust optimization scheme that formulated adversarially robust training as a convex optimization problem that can be solved using projected gradient descent (PGD). The attacker is allowed to create perturbations within a ball $S$ of radius $\epsilon$. PGD attempts to find a worst-case adversarial perturbation within $S$ that maximizes the loss on a given image, and then a classifier is trained on these perturbations:
%
 $   \min_\theta E_{(x,y) \sim D} \left [\max_{\delta \in S} L(\theta, x+\delta, y) \right].$
%
The resulting model has been shown to withstand against adversarial attacks belonging to $S$. However, training models in this manner leads to significant loss in natural accuracy \citep{tsipras2019robustness}.
%
Natural accuracy is an important characteristic for any model, including those trained to be adversarially robust. In practice, a model does not know whether it is being attacked, and if it is, it still does not know the strength of the attack. Therefore, it is preferable to have a system that adapts to different perturbation strengths while being agnostic to the perturbation itself. Our goal is to utilize ensemble schemes to improve natural accuracy while guaranteeing adversarial robustness at the same attack strength $\epsilon$. While ensemble schemes that use multiple classifiers and take a majority vote on a class prediction has been previously proposed \citep{tramer2018ensemble}, these approaches have been shown to be adversarially vulnerable \citep{he2017adversarial}.

{\bf Robust Ensembles:} Our key idea leverages the inverse relationship between the strength of the robustness guarantee (i.e., the size $\epsilon$ of the $L_p$ ball of attacks a model is exposed to during training) and natural accuracy \citep{tsipras2019robustness}. 
In particular, we point out that ensembling robust models can only improve adversarial accuracy. As a result, ensembling  models trained to withstand smaller $\epsilon$ attacks can withstand a much larger $\epsilon$ targeted attack. Furthermore, the inverse relationship of attack size and accuracy leads to significant improvements in natural accuracy. Consequently, we can optimize ensemble training to maximize natural accuracy for a desired targeted attack-level. 
%
%

{\bf Composing Robust and Natural Features:} We also explore a strategy based on fusing natural and robust models. 
These ideas are inspired from \citet{ilyas2019adversarial}, who reasoned that the existence of \textit{non-robust} and \textit{robust} features in images for natural accuracy loss in robust models. Non-robust features are human-imperceptible pixel patterns that still correlate well with class predictions. These result from the tendency of standard classifiers to use any available signal for generalization. Meanwhile, robust features are more or less human-interpretable patterns that can also generalize independently of non-robust features, though at significantly less accuracy than with both types of features used. Robust training removes the dependency of a model on non-robust features, and a higher attack strength strips out more non-robust features, while fusion leads to improved natural accuracy without suffering adversarial degradation. 

We propose an area-under-the-curve (AUC) metric to characterize adaptation to a continuous spectrum of attack strengths of adversarially robust models: 
%
 $$   AUC(\epsilon_{target}) = {1\over \epsilon_{target}}\int_0^{\epsilon_{target}} {\cal A}(\epsilon) d\epsilon.$$ Maximizing performance with respect to this metric would mean that a model would work well across the entire adversarial attack spectrum. 
%
%
Our results show that both our schemes result in significant improvement over robustly trained ResNet18 models on CIFAR-10 datasets both with respect to natural accuracy and AUC metric. 


\vspace{-1mm}
\section{Problem Formulation and Algorithm}
\vspace{-1mm}

For completeness we first describe the adversarial accuracy for robustly trained base-models. Let the input examples and labels belong to the product space ${\cal X}\times {\cal Y}$, and distributed according to the joint distribution denoted by $D$. Let $\theta$ denote the parameters of some model trained in some fashion on training data ${\cal D}=\{(x_i,y_i),\, i\in [n]\}$. Let $\theta(\alpha,\omega)$ be the (optimal) parameters of a robustly trained base model, with a random weight initialization $\omega$, which is supposed to withstand $\ell_2$ bounded test-time perturbations as large as $\eta$. Training a model robust to perturbations of size $\alpha$ follows \citet{madry2018towards}, where PGD is used to train on ${\cal D}$.

Conventionally, models are trained to withstand adversarial attacks of size $\eta$, and evaluated to validate attacks of this size, in this work, we are interested in its performance for other attack sizes. For instance, we are often interested in the natural accuracy of a model. Thus, we often allow for test-time perturbations, $0\leq \epsilon$, which can take on any non-negative value. As such our objective is to develop a perturbation-agnostic characterization of a model's performance. We denote the accuracy of a model, $\theta$ with respect to these parameters and loss function $L(\theta,x,y)$ as 
\begin{equation} \label{e.accuracy}
\rho(\theta,\epsilon) = \E \left [\max_{\|\delta\|\leq \epsilon} L(\theta,x+\delta,y) \right ]
\end{equation}
where the expectation is with respect to test-data.
The expected performance of a randomly initialized robustly trained model for attack size $\epsilon$ is the average over $K$ different random initializations, $\omega_1,\omega_2,\ldots, \omega_K$ and is denoted as $$\bar{\rho}(\eta,\epsilon) = \E_{\omega} [\rho(\theta(\eta,\omega),\epsilon)] \triangleq {1\over K}\sum_{j \in [K]} \rho(\theta(\eta,\omega_j),\epsilon).$$

\noindent {\bf Ensembling.}
In the previous scheme, a single base-model $\theta(\eta,\omega)$ is offered to the attacker, and the attacker is allowed to choose the worst perturbation for this model. In our random ensembling scheme, our model is a weighted average of many random base models, and the attacker must attack the average model rather than any model in isolation. 

In particular, let us now consider a collection of $K$ robust models trained with different random initializations $\Omega=\{\omega_1,\omega_2 \ldots \omega_K\}$, where each, $\omega_j$ accounts for different initializations in our training scheme. For instance, we could choose $\omega_j$'s independently from a multivariate Gaussian distribution, and this corresponds to independent random initializations of the model parameters. Although, one can experiment with different initialization strategies, and optimize over it, we only consider initializations drawn from a fixed distribution. Correspondingly we obtain $K$ robust models, $\theta(\alpha,\omega_j)$, which are each robust models that are trained for attacks of size $\alpha$. We can weight these models with non-negative weights $w_1,\,w_2,\ldots,w_K$ that sum upto one. Let $\bw$ denote the weight vector $(w_1,w_2,\ldots,w_K)$, which formally take values in a K-dimensional simplex $\Delta^K$. The ensemble model $\bar{\theta}(\alpha,\bw) = \sum_{j \in [K]} w_j \theta(\alpha,\omega_j)$. As a consequence, we are now faced with a different accuracy expression, namely, 
\begin{equation} \label{e.accuracy_ensemble}
\rho(\bar{\theta}(\alpha,\bw),\epsilon) = \E \left [\max_{\|\delta\|\leq \epsilon} L\left(\sum_{j \in [K]} w_j \theta(\alpha,\omega_j),x+\delta,y\right ) \right ]
\end{equation}
We note that whenever the loss function is convex (which is typically the case), it follows from Jensen's inequality that,
\begin{align*}
\min_{\bw \in \Delta^K}\rho(\bar{\theta}(\alpha,\bw),\epsilon) &\leq \min_{\bw\in \Delta^K} \E_D \left [\max_{\|\delta\|\leq \epsilon} \sum_j w_j L(\theta(\alpha,\omega_j),x+\delta,y) \right ] \\& \leq {1 \over K}\sum_{j\in [K]} \E_D \left [\max_{\|\delta\|\leq \epsilon} L(\theta(\alpha,\omega_j),x+\delta,y) \right ] = \bar{\rho}(\alpha,\epsilon)    
\end{align*}
{\bf Remark.}
Thus, what is clear is that {\it adversarial loss of a random ensemble is no worse than the average adversarial loss of a robust base model} trained with different initializations. 

{\bf Optimizing Natural Accuracy for a Target Adversarial Perturbation.}
These comments leads us to the following insight. Adversarial robustness is monotonic in the size of the perturbation, namely, $\bar{\rho}(\epsilon_1,\epsilon_1) \leq \bar{\rho}(\epsilon_2,\epsilon_2)$ for $\epsilon_1 \leq \epsilon_2$. Furthermore, it is generally true that the natural accuracy of an adversarially trained model for level $\epsilon$ decreases with $\epsilon$, namely, $\bar{\rho}(\epsilon_1,0) \leq \bar{\rho}(\epsilon_2,0)$ for $\epsilon_1 \leq \epsilon_2$. As a result, we propose strategies that combine models to achieve higher natural accuracy than a single robust base model, while suffering no additional adversarial degradation relative to the base model. We discuss two schemes to optimize natural accuracy below.

{\bf Averaging Randomly Initialized Robust Models.}
One way to do so is averaging of randomly initialized robust models. Here we train a collection of $K$ randomly initialized robust models at a perturbation level $\alpha$ which is significantly smaller than $\epsilon_{target}$. Specifically, our general objective can be framed as maximizing natural accuracy subject to no degradation in adversarial robustness:
\begin{align*}
    \min_{\alpha,\bw} \rho(\bar{\theta}(\alpha,\bw),0),\,\,\,
    \mbox{s.t.}\,\,\, \rho(\bar{\theta}(\alpha,\bw),\epsilon_{target}) \leq \bar{\rho}(\epsilon_{target},\epsilon_{target})
\end{align*}
We consider a simpler scheme here. We set weights to be uniform across all models. We then minimize the perturbation size $\alpha$, such that the adversarial error at target, $\epsilon_{target}$ is no worse than that of the base model. In practice we can validate this empirically on held-out validation data. 

{\bf Meta-Composites of Natural and Robust Models.}
It has been argued recently that robustness is a feature, and represents intrinsic and invariant properties of an object. Nevertheless, training with robust features typically result in significant loss in natural accuracy. Our goal here is to compose robust and natural features to form composite models, $\kappa(\theta,\theta')$ that lead to best of both worlds, namely, maintaining adversarial accuracy while significantly improving upon natural accuracy. To do so we train two models, one robust, $\theta(\epsilon,\omega)$ at say $\epsilon \Omega(\epsilon_{target})$, and the other, $\theta(\alpha,\omega')$ at $\alpha \approx 0$, where $\omega,\omega'$ are random initializations. We fuse information from both models by augmenting the penultimate layers of the two networks, one robust and the other natural, and perform PGD training only on the last layer to withstand an attack level $\epsilon$. This results in a composite model, $\kappa(\theta(\epsilon,\omega),\theta(\alpha,\omega')$. We then train a collection of such composites based on $K$ different random initializations and construct ensembles of such models by averaging their outputs as in the previous section. This results in 
$\bar{\kappa}(\epsilon,\alpha) =\sum_{j\in [K]}\kappa \left (\theta(\epsilon,\omega_j),\theta(\alpha,\omega'_j)\right )$. 
Our goal is to optimize $\epsilon$ and $\alpha$ so as to maintain target adversarial robustness while maximizing natural accuracy.

\vspace{-1mm}
\section{Results and Conclusions}
\vspace{-1mm}

{\bf Experimental Setup.}
We describe the train and test environment for our experiments. All base robust models were trained using PGD \citep{madry2018towards} for ResNet18 with the attack strength $\epsilon$ specified in the tables below and default attack steps. Models were trained (50,000 samples) and tested (10,000 samples) on the CIFAR10 dataset. Results reported are the best configurations found over 5 runs. 

{\bf Ensemble Accuracy.}
Table~\ref{overview} tabulates natural accuracy and adversarial robustness at given values of $\epsilon$ as well as the AUC($0.5$), namely, the area-under-the curve for targeted attacks below $\ell_2$ perturbations bounded by $0.5$.

We show that with eight robust models trained at $\epsilon = 0.22$, we can achieve the same robust accuracy as a single model trained at $\epsilon = 0.5$. A fundamental advantage of doing so is that each of the robust models now have substantially higher natural accuracy. This is a 6.4\% increase in natural accuracy compared to a single model at $\epsilon = 0.5$. We additionally note that the robustly averaged model has a higher adversarial accuracy before it converges with the single robust model when evaluating at $\epsilon = 0.5$. This gives it a higher AUC than the single robust model. Trivially, we also see that the ensemble has a better robust accuracy than a single natural model and a weakly trained base model.

A second point to observe is that our robust ensembles improve upon natural accuracy as well as adversarial accuracy. This is as expected because in Table 2 we note that the accuracy of the ensemble trained to withstand attacks of size $\alpha$, when ensembled is equivalent to a single robust model trained to withstand attacks of size $\epsilon=2\alpha$, namely, $\rho(\bar{\theta}, \alpha) > \bar{\rho}(\theta, 2\alpha)$.  
This implies that natural accuracy, i.e. a scenario where $\epsilon = 0$, should also improve with ensembling: $\rho(\bar{\theta}, 0) > \bar{\rho}(\theta, 0)$. Our results show that this is indeed the case. We also notice in Table~2 that natural accuracy saturates after ensembling a small number of models, implying most gains are early.

Using two models, a strong and a weak model, the single composite delivers better natural accuracy than a single robust model, but compromises slightly in robustness. However, combining two composites in a weighted average scheme, we do not compromise any robustness. The accuracy degradation curve over $\epsilon$ is less steep than robust averaging, even with fewer models used, and this is reflected by a stronger AUC metric value.

\begin{table}[]
\small
\label{overview}
\begin{center}
\begin{tabular}{|l|l|l|l|l|l|}
\hline
& $\epsilon=0$ & $\epsilon = 0.22$ & $\epsilon = 0.35$ & $\epsilon = 0.5$ & AUC($0.5$)     \\ \hline
Natural model& 94.64            & 13.5& 2.57& 0.44 &  0.067 \\ \hline
\begin{tabular}[c]{@{}l@{}}Robust Model\\ $\epsilon = 0.22$
\end{tabular}                                      
& 92.86& 81.24 & 71.71& 59.47& 0.722 \\ \hline
\begin{tabular}[c]{@{}l@{}}Robust Model\\ $\epsilon=0.5$\end{tabular}
& 88.30& 81.31& 78.11& 68.73&0.767  \\ \hline
\begin{tabular}[c]{@{}l@{}}8-Robust\\ $\epsilon = 0.22$\end{tabular}
& 93.95            & 83.95  & 78.92  & 68.79 & 0.781  \\ \hline
\begin{tabular}[c]{@{}l@{}}1-composite\\ $\epsilon = 0.4, \alpha=0.05$ \\ trained at $\epsilon = 0.4$\end{tabular} & 91.36& 83.22 & 76.62 & 67.95 & 0.769        \\ \hline
\begin{tabular}[c]{@{}l@{}}2X 1-composite \\ weighted average\tablefootnote{One model comprised of $\epsilon = 0.4, \alpha=0.05$ trained at $\epsilon = 0.4$ and weighted at 90\%, second model comprised of $\epsilon = 0.3, \alpha=0.05$ trained at $\epsilon = 0.3$ and weighted 10\%.}\end{tabular} & 91.61& 84.15 & 78.01 & 69.99 & 0.783        \\ \hline
\end{tabular}
\end{center}
\caption{\small{Results illustrate comparisons between single ResNet18 model trained to withstand targeted attacks of $\ell_2$ ball of size $\epsilon$ against our random ensembling schemes. Our ensembling schemes uniformly improve over attacks of different sizes as well as achieve similar performance at the targeted $\epsilon$-level of the robust model. A key insight is that ensembling schemes, whereby each model is trained for a smaller value $\alpha$ can withstand larger attacks when ensembled.}}
\end{table}
\begin{table}[]
\small
\begin{center}
\begin{tabular}{|l|l|l|}
\hline
\#Models & \begin{tabular}[c]{@{}l@{}}Nat. \\ Acc.\end{tabular} & \begin{tabular}[c]{@{}l@{}}Adv.\\ Acc.\\ $\epsilon = 0.5$\end{tabular} \\ \hline
1        & 88.3                                                           & 68.73                                                                           \\ \hline
2        & 88.92                                                          & 71.19                                                                           \\ \hline
4        & 89.07                                                          & 72.53                                                                           \\ \hline
8        & 89.36                                                          & 73.08                                                                           \\ \hline
12       & 89.28                                                          & 73.34                                                                           \\ \hline
16       & 89.18                                                          & 73.37                                                                           \\ \hline
\end{tabular}
%
\quad \quad \quad \quad 
\begin{tabular}{|l|l|}
\hline
\begin{tabular}[c]{@{}l@{}}$\alpha$ Ensemble\\ 8 models\end{tabular} & \begin{tabular}[c]{@{}l@{}} Single Robust\\ Equivalent ($\epsilon$)\end{tabular} \\ \hline
0.22                                                                        & 0.5                                                                    \\ \hline
0.25                                                                        & 0.59                                                                   \\ \hline
0.3                                                                         & 0.67                                                                   \\ \hline
0.4                                                                         & 0.75                                                                   \\ \hline
0.5                                                                         & 0.88                                                                   \\ \hline
\end{tabular}
\end{center}
\caption{\small{(Left) Table illustrates improvements in natural and adversarial accuracy with larger ensembles. Most gains are realized with 4-8 models. (Right) Illustrates equivalence between single robust model and a 8-robust ensemble. For a robust model trained to withstand different levels of targeted attacks $\epsilon$, we notice that for 8-model ensemble, we only need to ensure that each model in the configuration withstands attacks that are approximately half of $\epsilon$ without incurring additional performance degradation.}}
\end{table}

\small
\bibliography{iclr2020_conference.bib}
\bibliographystyle{iclr2020_conference}

\end{document}